\documentclass[journal]{IEEEtran}
\usepackage{cite}
\usepackage{amsmath,amssymb,amsfonts}
\usepackage{enumitem}
\usepackage{multirow}
\usepackage{booktabs}
\usepackage{algpseudocode}
\usepackage{pgfplots}
\usepackage{textcomp}
\usepackage{xcolor}
\usepackage{color}
\usepackage{graphicx}
\usepackage{subcaption}
\usepackage{amssymb}
\usepackage{pifont}
\usepackage{amsmath}
\usepackage[ruled,linesnumbered,noresetcount]{algorithm2e}

\begin{document}

\title{Position and Orientation-Aware One-Shot Learning for Medical Action Recognition \\from Signal Data}

\author{Leiyu Xie,~\IEEEmembership{Student Member,~IEEE,}
        Yuxing Yang,~\IEEEmembership{Student Member,~IEEE,}
        Zeyu Fu,~\IEEEmembership{Member,~IEEE,}\\ 
        and Syed Mohsen Naqvi,~\IEEEmembership{Senior Member,~IEEE}
\thanks{Leiyu Xie, Yuxing Yang and Syed Mohsen Naqvi are with Intelligent Sensing and Communications Group, School of Engineering, Newcastle University,
NE1 7RU Newcastle upon Tyne, UK (e-mail: l.xie6@newcastle.ac.uk; y.yang60@newcastle.ac.uk; 
mohsen.naqvi@newcastle.ac.uk).}
\thanks{Zeyu Fu is with the Department of Computer Science, University of Exeter, UK (e-mail: z.fu@exeter.ac.uk).}
}

\maketitle

\begin{abstract}
In this work, we propose a position and orientation-aware one-shot learning framework for medical action recognition from signal data. The proposed framework comprises two stages and each stage includes signal-level image generation (SIG), cross-attention (CsA), dynamic time warping (DTW) modules and the information fusion between the proposed privacy-preserved position and orientation features. The proposed SIG method aims to transform the raw skeleton data into privacy-preserved features for training. The CsA module is developed to guide the network in reducing medical action recognition bias and more focusing on important human body parts for each specific action, aimed at addressing similar medical action related issues. Moreover, the DTW module is employed to minimize temporal mismatching between instances and further improve model performance. Furthermore, the proposed privacy-preserved orientation-level features are utilized to assist the position-level features in both of the two stages for enhancing medical action recognition performance. Extensive experimental results on the widely-used and well-known NTU RGB+D 60, NTU RGB+D 120, and PKU-MMD datasets all demonstrate the effectiveness of the proposed method, which outperforms the other state-of-the-art methods with general dataset partitioning by 2.7\%, 6.2\% and 4.1\%, respectively.
\end{abstract}


\IEEEpeerreviewmaketitle

\section{Introduction}
\IEEEPARstart{H}{uman} action recognition has played an important role in recent years for many applications, such as human-machine interaction, video surveillance, and the healthcare system \cite{zhang2018fusing, liu2020multi, li2023graph, shi2020skeleton, peng2023delving}. The major task of human action recognition is to accurately analyze and classify human behavior based on the given action sequences \cite{ahn2023star, duan2022revisiting, zou2023learning, trivedi2023psumnet, liu2023skeleton}. However, different from the general action recognition task, medical action recognition has many fewer samples. There are several reasons leading to this issue. Firstly, similar to other medical applications, privacy protection is one of the most challenging topics for medical action recognition \cite{xie2023one, xie2022privacy}. It includes safeguarding facial information, dressing information, and the surrounding background information. Secondly, medical actions such as coughing, staggering, falling, headache, vomiting and falling occur much less frequently than normal human actions in everyday life \cite{chou2018privacy}. Some of these medical actions may not even occur within a month. As a result, the number of samples available for analysis is significantly smaller compared to normal actions. In addition, there have been plenty of approaches proposed for human action recognition based on skeleton data, which serves to decrease the impact of dynamic illumination and mitigate the privacy leakage issues \cite{cai2021jolo,joze2020mmtm,li2020sgm,shahroudy2017deep,li20213d,duan2022revisiting,zhang2018fusing,song2022constructing}. 


\begin{figure}[t]
\centering
\includegraphics[width=8.9cm, height=6.9cm]{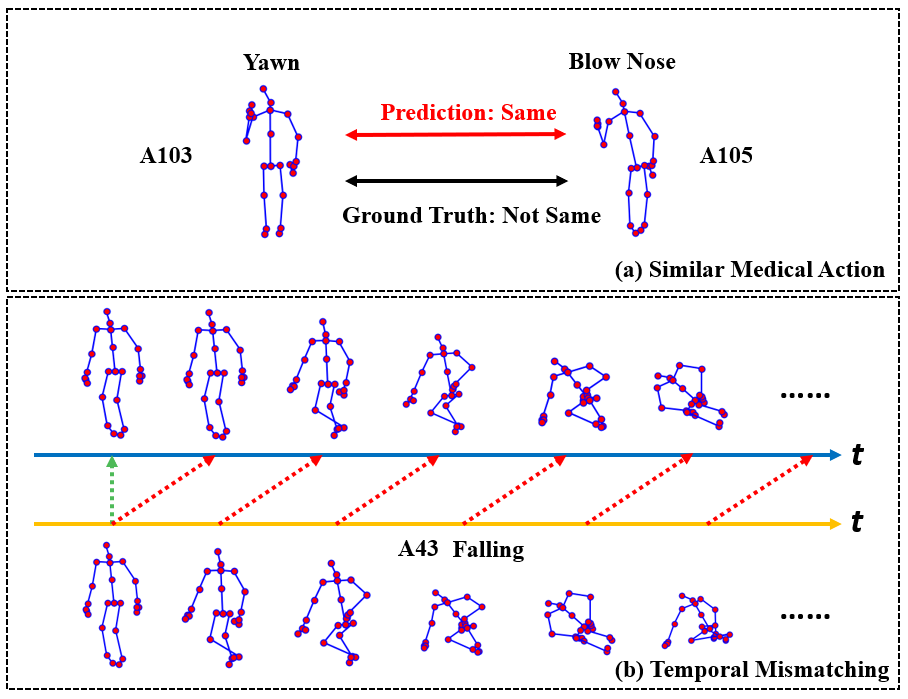}
\caption{The illustration of two primary limitations existing in the conventional one-shot learning framework: (a) similar medical actions, (b) temporal mismatching issues. The instances are from NTU RGB+D 120 dataset \cite{liu2019ntu}, detailed descriptions and labels are also provided.}
\centering
\end{figure}


With the increasing benefit of one-shot learning, the learning-with-limited-samples framework has contributed attractively successful results to medical action recognition based on human skeleton sequences \cite{peng2023delving, memmesheimer2022skeleton}. One-shot learning refers to feeding the model with a single instance for each category, accompanied by prior knowledge \cite{zou2020adaptation, han2022one}. Apart from the aforementioned issues, there are still two primary challenging limitations that hinder recognition performance in this learning framework: similar actions and temporal mismatching \cite{yin2023efficient, rachna2023real, liu2023novel, guo2023motion}, which are illustrated in detail in Fig. 1. Firstly, similar medical actions exist in real-life environments, such as headache and neck pain, putting on glasses and taking off glasses. Similar actions may confuse the network and lead to incorrect training directions since they have different importance for their landmarks \cite{ahn2023star}. This may also drastically degrade experimental performance when matching the support and query set, as the unimportant landmarks can have a noise-like effect during the matching process \cite{kang2023action}. Secondly, since the same labeled action sequences may have different action timing and temporal lengths, variations in the speed at which subjects demonstrate the same action can result in different lengths of action sequences \cite{ram2023enhanced}. This temporal mismatching issue will decrease the performance for the framework when matching the sequences from the support and query sets \cite{ma2022learn}.

\begin{figure}[t]
\centering
\includegraphics[width=8.6cm, height=5.9cm]{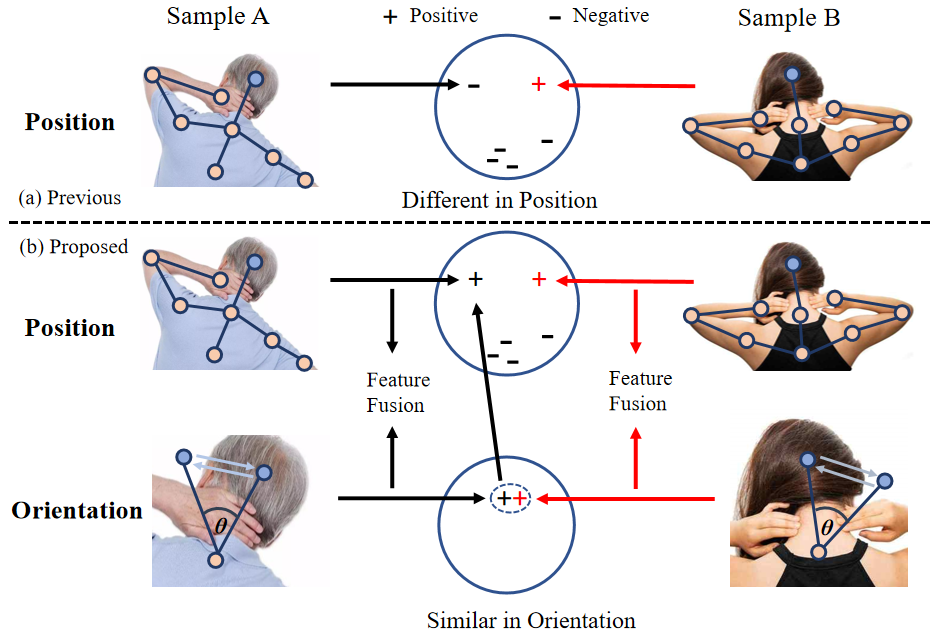}
\caption{Neck pain actions in position and orientation formats. (a) The previous approaches using position features predict them as different actions. (b) The proposed orientation features and assisted training method enhance the recognition performance and consider the samples as the same actions.}
\centering
\end{figure}

\begin{figure*}[htbp!]
\centering
\includegraphics[width=17.4cm, height=9.4cm]{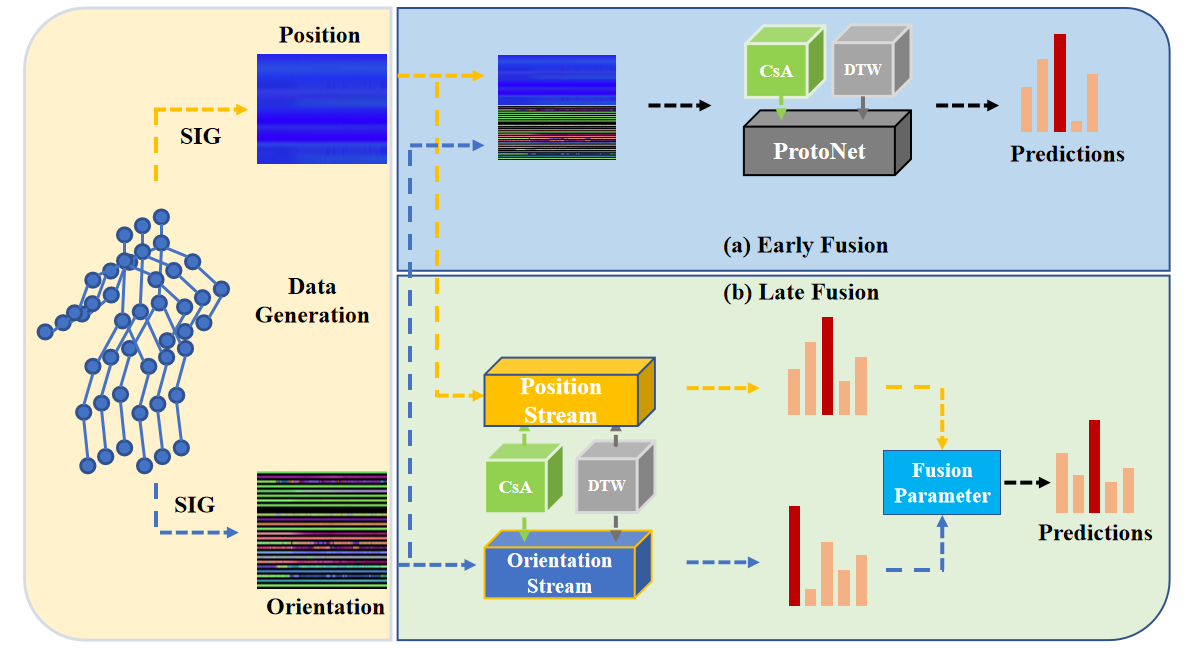}
\caption{The overall illustration of proposed POA-OSL. Raw skeleton sequences are first transformed into signal-level images by the SIG method and then fed to different fusion levels for medical action recognition. (a) The early fusion aims to concatenate position and orientation features as the final feature set and fed it into the proposed framework for medical action recognition. (b) Both CsA and DTW modules are employed in the late fusion. The probability scores from the multiple streams are calculated for the final predictions (best view this in the color version).}
\centering
\end{figure*}


%

Most previous works aim to address the aforementioned issues solely focused on applied feature vectors consisting solely of the coordinates of the 3D human landmarks, which could be considered as the position-level features of the skeleton data \cite{li2023spatial, lee2023mmts}. However, we argue that the orientation-level features can provide a more informative interpretation in the temporal dimension, which could assist the position-level feature to further discriminate similar medical actions and improve the recognition performance as shown in Fig. 2. 

Therefore, we propose a position and orientation-aware one-shot learning framework (POA-OSL) for medical action recognition in this work. The proposed framework consists of two stages and each stage contains three primary components: signal image generation (SIG), cross-attention (CsA) and dynamic time warping (DTW). In the SIG stage, the raw skeleton sequences are transformed into an image-based feature paradigm to further mitigate privacy leakage. Following with the proposed CsA module, which aims to guide the model to prioritize the more crucial landmarks from the data, which helps the model to discriminate similar actions and reduce the misclassification situation. The DTW module is designed to assist the model in aligning the temporal information of two instances during the matching stage, which helps alleviate the performance degradation caused by mismatched action timings. Moreover, the extracted orientation-level features are applied to fuse with the position-level features in both two stages for further discriminating similar actions, which investigates their complementary benefits for medical action recognition. The overview of the proposed medical action recognition framework is illustrated in Fig. 3. To summarize, the main contributions of our work are listed as follows:

\begin{enumerate}
    \item A novel feature transformation method which distilled the skeleton sequences for both position and orientation features is proposed for further preserving privacy and improving the recognition performance.
    \item A novel medical action recognition approach is proposed by conducting a feature cooperative training method within the one-shot learning framework.
    \item Cross-attention and dynamic time warping modules are applied for addressing similar medical actions and temporal mismatching issues to enhance the training process.
    \item Experimental results on NTU RGB+D 60 \cite{shahroudy2016ntu}, NTU RGB+D 120 \cite{liu2019ntu} and PKU-MMD \cite{liu2017pku} are provided to confirm improved performance over the state-of-the-art one-shot learning action recognition methods.
\end{enumerate}

Preliminary work has previously been presented in ICASSP 2023 \cite{xie2023one}, which was mainly to address the similar action and temporal mismatching issues in the medical action recognition task. In this work, we present the orientation-level feature extraction method for the first time. The previous work \cite{xie2023one} is extended to a multi-level feature extraction approach with a one-shot learning framework, including the novel proposed orientation-level assisted training approaches to perform medical action recognition. Extensive experiments on different benchmark datasets are conducted to compare the proposed method with the state-of-the-art works. Furthermore, the ablation study with different modules and different parameter analyses are also presented in this work.

\section{Related Work}
In this section, we will briefly introduce related works on human action recognition using skeleton data and one-shot learning, as well as analyze the difference between previous works and the proposed approach.

\subsection{Skeleton-based Human Action Recognition}

Human skeleton sequences often comprise multiple human body landmarks, which serve as representations of human actions. Utilizing skeleton sequences enhances performance robustness against background information and dynamic illumination when compared to previous approaches employing conventional RGB image data \cite{angelini20192d, 8656578}. As a result, the utilization of skeleton sequence data for human action recognition has garnered increasing attention and popularity, particularly with the advancement of data-driven deep learning methods. For instance, \cite{du2015hierarchical} introduced an LSTM-based approach to encode temporal information from skeleton sequences. These sequences are partitioned into five segments based on the human physical body structure and subsequently inputted into corresponding sub-networks.

With the advancement of graph neural networks, human action recognition has garnered even more attention due to its efficient modeling capabilities. In \cite{shi2019skeleton}, the authors proposed DGNN, which extracts joint and bone information based on the relationships within skeleton sequences. Moreover, they designed an adaptive topological structure for the graph to further enhance the performance of the model. For a deeper exploration of skeleton sequences, \cite{yu2017spatio} introduced ST-GCN, capable of extracting spatial and temporal information through graph convolutional network (GCN) blocks. They employed GCN to derive spatial features from human joints in the spatial dimension and extracted temporal features from adjacent frames using the temporal graph convolutional network model. MS-G3D \cite{liu2020disentangling} proposed a unified spatial-temporal graph convolutional operator to disentangle the significance of nodes in neighbouring contexts. \cite{chen2021channel} put forward CTR-GCN, a channel-wise topology refinement graph network, which applies shared topology features to all channels as a generic prior knowledge. While these methods primarily focused on model design, they did not fully address the privacy protection issue. Despite raw skeleton sequences mitigating much of the personal information, landmark localization still persists. Therefore, in this work, we aim to further tackle the privacy protection challenge by proposing the SIG approach that converts raw skeleton sequences into signal-level images.

\subsection{One-Shot Learning}
Differing from previous data-driven approaches, one-shot learning for action recognition tasks focuses on classifying the query set using only a single labeled instance for each class \cite{wang2022temporal, sabater2021one, hong2021video, li2022smam, zhu2023adaptive, chen2023part, wang2023neural, lee2023mmts}. This approach helps tackle the scarcity of labeled data due to high costs and privacy concerns. It also meets the demand for recognizing entirely new action categories in our daily lives. To address occlusion issues in one-shot action recognition, Trans4SOAR was proposed in \cite{peng2023delving}, aiming to mitigate the negative effects caused by occlusions. In \cite{zhu2023adaptive}, an approach was introduced to address the instability of representation resulting from insufficient generalized learning on invariant samples and noisy features. They presented an adaptive local component-aware GCN that calculates similarity based on aligned local embeddings of action segments. In \cite{chen2023part}, a part-aware prototypical representation was suggested to reduce the independence of body-level features in the spatial dimension. For challenging therapy scenarios, \cite{sabater2021one} introduced an anchor and target motion-based framework. Although the aforementioned methods have achieved state-of-the-art performance in one-shot action recognition, temporal information mismatching issue often exists between sequences from the same action class. Moreover, the sample features between the support and query sets are not shareable, which limits potential performance enhancement. To address these issues, cross-attention and dynamic time warping mechanisms are introduced into our proposed one-shot learning framework for medical action recognition.

\section{Proposed Methods}
\subsection{Preliminaries}
To address the aforementioned limitations in the one-shot medical action recognition, we propose a prototypical network-based \cite{snell2017prototypical} novel one-shot learning framework with an orientation-level assisted training method. The proposed algorithm will be demonstrated on the dataset $\mathcal{D}=\left\{\left(S_i, y_i\right)\right\}_{s=1}^{\mathcal{N}}$ which includes $\mathcal{N}$ skeleton sequences $S_{1},...S_{\mathcal{N}}$, with given labels $y_i \in\{1, \ldots, \mathcal{M}\}$. Two different human action features, $I_{j}$ and $I_{a}$ are extracted from $S_{i}$ with signal-level image representation for position and orientation features, respectively. We aim to train the extracted features $I_{j}$ and $I_{a}$ from dataset $\mathcal{D}$ to get the position feature representation $\Vec{x}_{j} = f_{\delta}(I_{j})$ and the orientation feature representation $\Vec{x}_{a} = f_{\delta}(I_{a})$. And utilizes $\Vec{x}_{j}$ and $\Vec{x}_{a}$ with the assisted training approaches by calculating the sequence distance based on the proposed metric learning framework for human medical action recognition.

\subsection{Signal Images Transformation}
\subsubsection{Position Feature Transformation}

\begin{figure}[h]
\centering
\includegraphics[width=6.9cm, height=5.6cm]{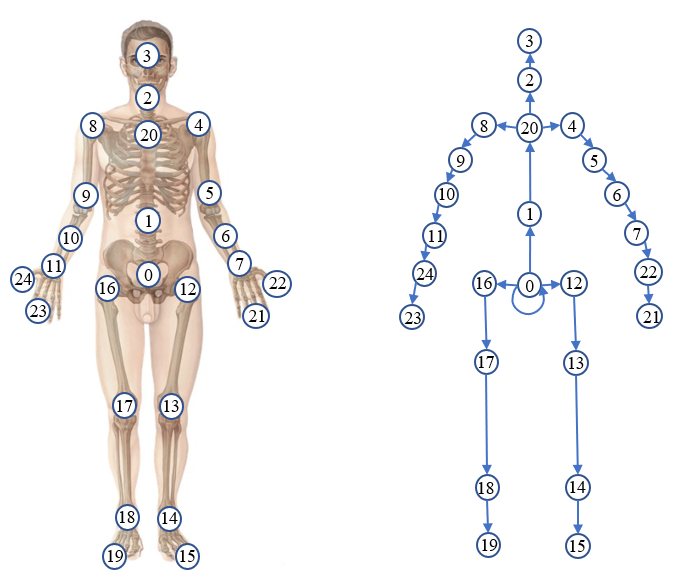}
\caption{The left sketch illustrates the position feature labels for each body part from NTU RGB+D 60, NTU RGB+D 120 and PKU-MMD. The right sketch shows the bone labels which are extracted for the orientation features and the directions of the bones are also demonstrated.}
\centering
\end{figure}

Based on the human physical structures, we manually design the human bones from the landmarks in the raw skeleton sequences. The customised landmark connections for the human bone information from the datasets are shown in Fig. 4. Different from most of the available skeleton-based approaches, in this work, the transformed signal level images are applied as the training data, the similar to \cite{xie2023one}. The raw skeleton sequence $S$ from NTU RGB+D 60, NTU RGB+D 120 and PKU-MMD datasets denote as $S = \mathbb{R} ^{X\times T\times V}$, where $X$ denotes the number of landmarks in each skeleton, $T$ denotes the temporal length of the sequence and $V$ denotes the 3D coordinates position of each landmark. Since the pixels of RGB image have three colour channels, $S$ could be transferred into image representation as $I_{j} \in \left \{ 0,1,...,255 \right \}^{H \times W \times 3}$. Hence, the total number of landmarks $X$ and the temporal length $T$ are transformed into the image height of $H$ and the image width $W$, respectively. To balance the impacts of the pixel values on the model performance, a normalization operation is also applied. A detailed illustration of the transformation process is shown in Fig. 5.


\begin{figure}[h]
\centering
\includegraphics[width=7.4cm, height=6.3cm]{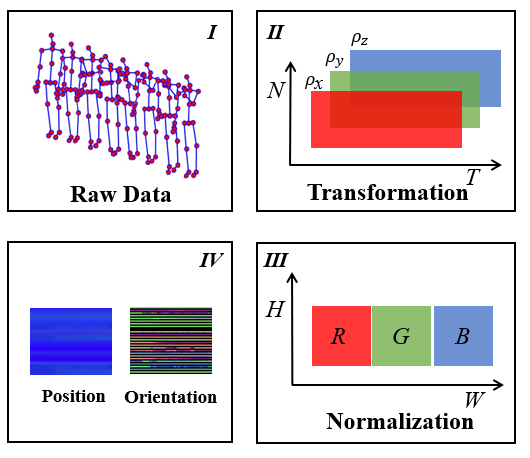}
\caption{The illustration of the proposed SIG method for different features. For the position features, the 3D coordinates position $V$ are transformed into $\rho_{x}, \rho_{y}, \rho_{z}$. For the orientation features, $\theta_{x}, \theta_{y}, \theta_{z}$ are transformed into $\rho_{x}, \rho_{y}, \rho_{z}$, respectively. The transformed features of position and orientation are shown in step \textit{\uppercase\expandafter{\romannumeral4}}.}
\centering
\end{figure}

\subsubsection{Orientation Feature Transformation}
In order to obtain different types of human action features from the raw skeleton sequences, the angles of the human bones are extracted as the orientation features for assisting the training process. 


Since the provided data is in 3D coordinates, in order to facilitate the acquisition of orientation information, we establish three normal vectors based on a coordinate system. By calculating the orientation information between each bone with the three normal vectors, we obtain human orientation features that are complementary to the position information for recognition accuracy. The orientation features with the three coordinate system planes are formulated as follows:
\begin{equation}
    \theta_{x} = \arccos{\frac{|\mathbf{v}_{x} \cdot \mathbf{v}_{b}|}{|\mathbf{v}_{x}| \cdot|\mathbf{v}_{b}|}}
\end{equation}
\begin{equation}
    \theta_{y} = \arccos{\frac{|\mathbf{v}_{y} \cdot \mathbf{v}_{b}|}{|\mathbf{v}_{y}| \cdot|\mathbf{v}_{b}|}}
\end{equation}
\begin{equation}
    \theta_{z} = \arccos{\frac{|\mathbf{v}_{z} \cdot \mathbf{v}_{b}|}{|\mathbf{v}_{z}| \cdot|\mathbf{v}_{b}|}}
\end{equation}
where $\mathbf{v}_{b}$ is the bone vector, $\mathbf{v}_{x}, \mathbf{v}_{y}, \mathbf{v}_{z}$ are the normal vectors, respectively and $\mathrm{\arccos(\cdot)}$ indicates the inverse cosine function for calculating the orientation features. The same as the position features, the orientation information will be also transferred into $I_{a} \in \left \{ 0,1,...,255 \right \}^{H \times W \times 3}$. $\theta_{x}, \theta_{y}, \theta_{z}$ will be calculated as the values of three colour channels for the pixels after the normalization. Similar to the position feature, temporal length $T$ and the number of landmarks $X$ are transferred into the height $H$ and the width $W$ of the signal level images, respectively. The detailed illustration is shown in Fig. 5.

\begin{figure*}[t]
\centering
\includegraphics[width=17.4cm, height=9.3cm]{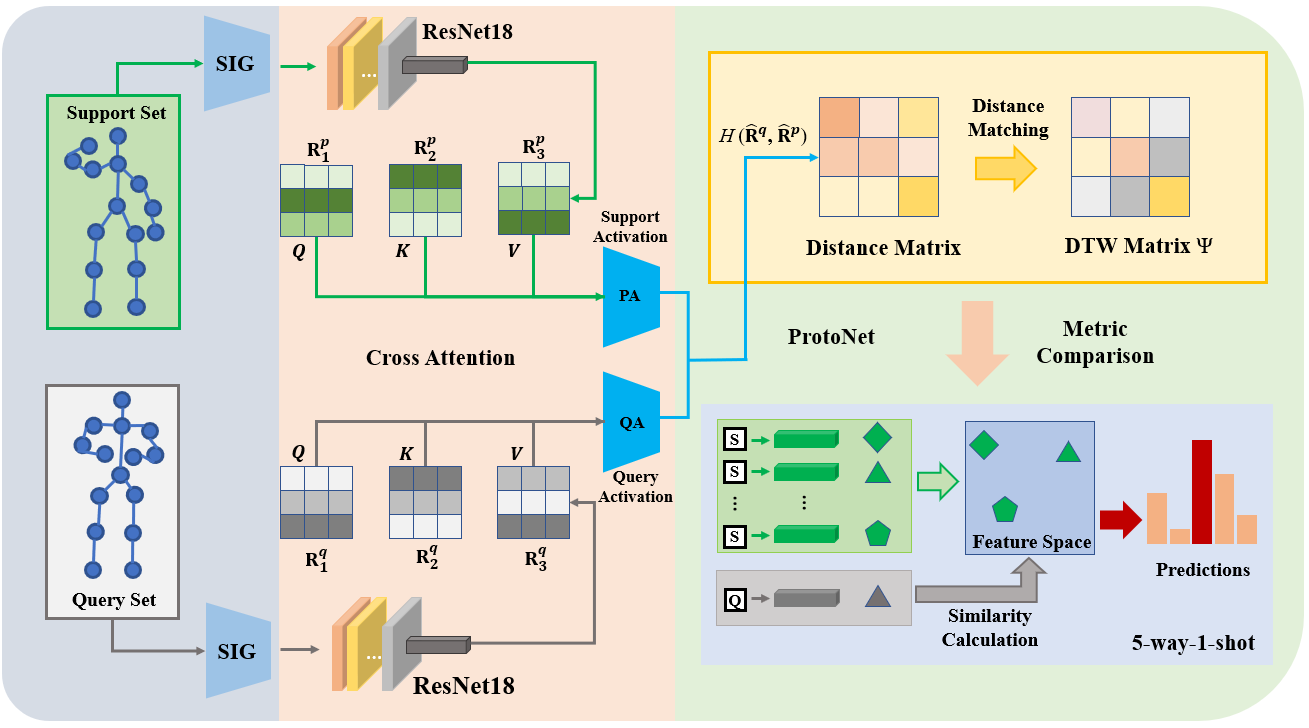}
\caption{The single-stream illustration of the proposed one-shot learning approach which contains the SIG, CsA and DTW modules. The data transformation (SIG) module first transforms the input skeleton sequences into signal-level images before being fed into the ResNet18 encoder for feature extraction. The encoded features from both support and query sets are fused via a cross-attention module for metric learning in the ProtoNet framework \cite{snell2017prototypical}. Dynamic time warping (DTW) module is exploited to address the temporal information mismatching issue which could be obtained via (6) and (7). The vectors from the support and query set are finally mapped to the feature space for similarity calculation and yield the final results.}
\centering
\end{figure*}

\subsection{Cross Attention Mechanism}
After transferring the skeleton sequence into the signal-level representation, a cross-attention module between the support set and query set is exploited in the proposed framework. In previous cross-attention approaches, typically, only one of the two modules involved in the computation was focused. The aim of cross-attention is to guide the network to give more attention to the important parts rather than the other parts. This mechanism could decrease the difficulties in discriminating against similar medical actions. This is because the spatial relationship between the different parts of the human body plays an important role in distinguishing similar actions. For example, coughing is similar to neck pain, the difference between them is that the head landmark of coughing is more important than the neck landmark for the neck pain action. Furthermore, the representation of the support set should adaptively change the importance of each landmark according to the representation of the query set or vice versa. The transformed representations of query set $\hat{\mathbf{R}}^q$ and support set $\hat{\mathbf{R}}^p$ are formulated as follows:
\begin{equation}
    \hat{\mathbf{R}}^q=\mathrm{S}\left(\frac{\mathbf{M}_1^q \mathbf{R}^q \cdot\left[\mathbf{M}_2^q \mathbf{R}^p\right]^\mathbf{T}}{\sqrt{d}}\right) \mathbf{M}_3^q \mathbf{R}^q
\end{equation}

\begin{equation}
    \hat{\mathbf{R}}^p=\mathrm{S}\left(\frac{\mathbf{M}_1^p \mathbf{R}^p \cdot\left[\mathbf{M}_2^p \mathbf{R}^q\right]^\mathbf{T}}{\sqrt{d}}\right) \mathbf{M}_3^p \mathbf{R}^p
\end{equation}
where the representation of the query set and support set could be formulated as $\mathbf{R}^{\mathit{q}} $ and $\mathbf{R}^{\mathit{p}} $, respectively. And $\mathrm{S} (\cdot)$ denotes the Softmax function which is applied for calculating the weights of different human body parts. $\mathbf{M}_1, \mathbf{M}_2, \mathbf{M}_3$ are the transformation matrices, which contain the trainable parameters. $\mathbf{T}$ denotes the transpose matrix operation and $d$ indicates the dimension of the data feature representation. 
\begin{equation}
    H\left(\hat{\mathbf{R}}^q, \hat{\mathbf{R}}^p\right)=\left\|\hat{\mathbf{R}}^q-\hat{\mathbf{R}}^p\right\|_F
\end{equation}
where $H$ denotes the distance between the two samples from the support and query sets, respectively. This will be applied in prototypical networks \cite{snell2017prototypical} for classifying different human actions. And $\|\cdot\|_F$ indicates the Frobenius normalization which is a kind of specific $L_{2}$ regularization between the matrices. In (4) and (5), $\mathbf{M}_2^q \mathbf{R}^p$ and $\mathbf{M}_2^p \mathbf{R}^q$ allow the model to interact with the feature information from different perspectives. The model can improve its accuracy in recognizing and classifying similar actions by incorporating its attention mechanism to further pay attention to the important landmarks through the interactive information between the support and query sets. This approach aims to minimize the misclassifications of similar actions. By applying higher attention weights to the more crucial parts of the human body, the model enhances its ability to distinguish between similar actions and ultimately enhances its overall accuracy.

\subsection{Dynamic Time Warping}
There exist several factors (e.g. different experimental subjects, speed, duration of the recording and action timing) that result in the temporal information mismatching issue between the support set and query set actions. For the instance, in the support set $P=\left\{\hat{\mathbf{R}}_1^p, \hat{\mathbf{R}}_2^p, \cdots \hat{\mathbf{R}}_m^p\right\}$ and the query set $Q=\left\{\hat{\mathbf{R}}_1^q, \hat{\mathbf{R}}_2^q, \cdots \hat{\mathbf{R}}_m^q\right\}$, $m$ is the length of the resized signal image. The mismatching issue will directly affect the Euclidean distance calculation between $\hat{\mathbf{R}}_i^p$ and $\hat{\mathbf{R}}_i^q$ and decrease the classification performance. To address this issue, we exploit the most popular temporal information alignment approach, which is the dynamic time warping approach from \cite{ma2022learn} to address this issue.

\begin{equation}
    \Psi(i, j)=E(i, j)+\min \{\Psi(i-\tau, j-\tau), \Psi(i-\tau, j), \Psi(i, j-\tau)\}
\end{equation}
where $\Psi(i, j)$ indicates the cumulative distance between the $i$-th frame from the query set and the $j$-th frame from the support set. $\tau$ is the time mismatch hyperparameter. Each element in $\textit{E(i, j)}$ is generated according to (6). In practice, each pair of instances is correspondingly associated with each other to compute a correlation distance response map. The instances could be aligned by measuring the similarity between them since DTW allows for the non-linear mapping of the temporal dimension. The $\hat{P}$ and $\hat{Q}$ will be updated from support set $P$ and the query set $Q$ after the DTW module, respectively. The operation $\mathrm{min}$ could be replaced by another operator to get a differentiable distance calculation. The single-stream illustration of the proposed one-shot learning approach is presented in Fig. 6.


\subsection{Orientation-level Feature Assisted Training}
Due to the raw data utilized, which only consists of coordinate position information for human landmarks, relying solely on a single-level feature for action recognition is insufficient to capture comprehensive action characteristics. Therefore, we propose an orientation-level assisted training approach to enhance the model performance in different stages, which primarily consists of both early fusion and late fusion methods. 

\subsubsection{Early Fusion}
In this context, early fusion actually enhances the representation of human action from multiple features. It refers to merging both the position and orientation signal-level images at an early stage of the proposed framework. It involves combining the complementary feature information and increasing the correlation between the position and orientation features before any further processing stage begins. 
The commonly used position-level information of the skeleton data only contains 2D or 3D coordinates of the landmarks. Nevertheless, the angles of the human bones, which are regarded as orientation-level features are naturally complementary to the position-level information. Typically, the movement differences in directions are supportive and discriminative for action recognition.

After the signal image transformation, position and orientation features are merged as one single image representation. However, the temporal dimensions of each raw skeleton sequence are different, to ensure consistent signal image size after transformation, we resize the feature image with the customised image resolutions. Early fusion can benefit from cooperative learning and feature sharing across both the position- and orientation-level information, potentially leading to improved performance. Moreover, the computational complexity could be simplified and straightforward to implement and interpret due to the increased feature dimensions. A detailed illustration of early fusion is shown in Fig. 3 (a).


\subsubsection{Late Fusion}
Since the early fusion merges position and orientation information at the initial stage of the proposed framework. The fused features may be sensitive to noise and outliers, if it includes potentially noisy information in the individual feature. Furthermore, due to the various complexity of the human action dataset, late fusion is needed for medical action recognition. Without the correlations between the position and orientation features, the prediction probability values from multiple streams will be weight averaged and obtained for the fusion recognition task. We demonstrate the pipeline of late fusion in Fig. 3 (b) for further explanation. Both the position and orientation features are trained in a multiple streams framework, and the late fusion matching prediction is calculated as:

\begin{equation}
    \mathcal{P}_{m} = \mathcal{P}_{j} + \alpha\mathcal{P}_{a}
\end{equation}
where $\mathcal{P}_{j}$ and $\mathcal{P}_{a}$ are the position feature prediction values and orientation feature prediction values from the sub-streams, respectively. $\alpha$ indicates the importance of weight and is set as 1 as the default. A detailed value analysis of the weight importance setting will be discussed in the experiment session.

\subsection{Training Objectives}
For the training set, there are $N$ classes with $K$ labelled support actions in each class. As the definition in \cite{snell2017prototypical}, each prototype indicates the mean vector of the support embedding points which belong to its categories. The prototypical representation of each instance could be formulated as follows:

\begin{equation}
    C_k=\frac{1}{K} \sum_{\left(\mathcal{A}_i^p, y_i^p\right)} f_\phi\left(\mathcal{A}_i^p\right) \times \mathbb{F}\left(y_i^p=k\right)
\end{equation}
where $\mathcal{A}_i^p$ is the actions in the support set and $\mathbb{F}$ is the indicator function. $f_\phi(\cdot)$ denotes the action encoder for the learnable parameter $\phi$ and $y_i^p$ denotes the ground truth label for the $i$-th instance from the support set. The prediction distribution of the actions from the query set is defined as follows:

\begin{equation}
    p_\phi\left(y=k \mid \mathcal{A}^q\right)=\frac{\exp \left(-\operatorname{dis}\left(f_\phi\left(\mathcal{A}^q\right), C_k\right)\right)}{\sum_{k^{\prime}} \exp \left(-\operatorname{dis}\left(f_\phi\left(\mathcal{A}^q\right), C_{k^{\prime}}\right)\right)}
\end{equation}
where $\mathcal{A}^q$ denotes the query action. And $\exp(\cdot)$ is the exponential function and $\mathrm{dis}(\cdot)$ is the distance function. Training episodes are generated by randomly selecting a subset of classes from the full-size training set. A subset of the instances from each class will be selected as the support set and the remaining instances as the query points to calculate the instance distance for classification. The matching loss $\mathcal{L}_{\text {m}}$ is derived as follows. 

\begin{equation}
    \mathcal{L}_{d}=-\frac{1}{B \times m} \sum_b^B \sum_i^m\left\|\mathbf{U_{i}}\right\|
\end{equation}

\begin{equation}
    \mathcal{L}_{\text {m}}=-\frac{1}{N^q} \sum_i^{N^q} \log p_\phi\left(\hat{y}_i=y_i \mid \mathcal{A}_i^q\right)+\lambda \mathcal{L}_{\text {d}}
\end{equation}
where $\mathcal{L}_{d}$ is the disentanglement loss function for decreasing the linear dependence between the skeleton key points. $B$ denotes the batch size and $m$ is the length of the resized image. $\mathbf{U_{i} \in \mathbb{R}^{N \times d} }$ denotes the updated image presentation for the $i$-th instance from $\hat{P}$ and $\hat{Q}$. The model training proceeds to minimise the negative log probability of the ground truth $y_{i}$ with the optimizer. $N^{q}$ indicates the number of query actions and $\hat{y}_i$ is predicted label of the $i$-th instance. $\left\|\cdot\right\|$ indicates the paradigm function and $\lambda$ is the weight hyperparameter of $\mathcal{L}_{d}$ for obtain the optimal model performance.

\section{Experimental Results}
In this section, we conduct the experiments on three public datasets which are most widely used and well-known for action recognition tasks including NTU RGB+D 60 \cite{shahroudy2016ntu}, NTU RGB+D 120 \cite{liu2019ntu} and PKU-MMD \cite{liu2017pku}. The quantitative results are also presented to compare with the other state-of-the-art one-shot learning methods for human action recognition. Moreover, we design experiments for specific medical actions analyzing as well as the result visualisation. Furthermore, we carry out ablation studies to demonstrate the effectiveness of transformed features, the proposed cross attention and dynamic time warping modules. Finally, the experiments for different parameter settings are presented.

\subsection{Datasets}
This section briefly introduces the three selected public and available datasets for the evaluations of the proposed POA-OSL. These datasets are distinctively and randomly divided into training (80\%), validation (10\%) and testing (10\%) sets.
The instances in the testing and validation sets are entirely individual with the training set.

\subsubsection{PKU-MMD}
PKU-MMD dataset is a 3D large-scale dataset that contains 1076 long skeleton sequences in 51 action classes. It is recorded from 66 subjects in 3 different camera views. There are over 20,000 instances provided with multi-modal data, including RGB, depth, infrared radiation and skeleton sequences. We select 5 medical-related actions as the testing set which contains, falling, backache, heart pain, headache and neck pain (A11, A42, A43, A44, A45). There are 5 randomly selected classes applied as the validation set and 41 classes of actions for the training set.

\subsubsection{NTU RGB+D 60}
NTU RGB+D 60 is a 3D large-scale human action dataset which provides skeleton data sequences. The skeleton data sequences consist of 56,880 instances for 60 types of human actions. These sequences are recorded from 40 different subjects with 17 different scene conditions, and each subject provides 25 pose landmarks. For evaluation of medical action recognition, 6 actions with medical conditions are considered for testing, they are cough, falling, headache, chest pain, back pain and neck pain (A41, A43, A44, A45, A46, A47).

\subsubsection{NTU RGB+D 120}
NTU RGB+D 120 dataset is an extended version of NTU RGB+D 60 dataset. It contains 120 types of human actions recorded from 106 subjects in 155 different scene conditions and each subject also provides 25 pose landmarks. There are over 114,000 skeleton sequences including daily, interaction and medical-related actions. For medical action recognition, 12 actions with medical conditions are selected as testing set, they are cough, staggering, falling, headache, chest pain, back pain, neck pain, vomiting, fan self, yawn, stretch oneself and blow nose (A41, A42, A43, A44, A45, A46, A47, A48, A49, A103, A104, A105). There are 12 classes selected as the validation set and the rest of the actions are set as the training set which contains 96 classes. There are 5 classes are randomly selected as the validation set and the rest 41 classes of actions are denoted as the training set.

\subsection{Evaluation Metrics}
We follow the most widely used and well-known standard evaluation metric from the SOTA works to examine the proposed framework, which is called Top-1 accuracy. It refers to the percentage of times that model correctly predicts the highest-probability class for the specific input sequence. For all the experiments, Top-1 accuracy is selected for comparing the performance with the SOTA approaches.

\subsection{Implementation Details}
For all the datasets, we use ResNet18\cite{he2016deep} as the backbone to encode the proposed image-level pose feature, which is the most widely used network for image-processing tasks. Adam optimizer\cite{kingma2014adam} is applied to the experiments with an initial learning rate of 0.001 with a decay of 0.5. The random seed is chosen as 7 and ProtoNet\cite{snell2017prototypical} is selected for classifying the actions by calculating the distances between the instances. For all three datasets, 1000 randomly selected iterations are applied for the training stage and 500 randomly selected iterations are applied for both the validation and testing stage. The parameter of DTW is empirically set as 1 and the loss parameter $\lambda$ is set to default as 0.1. Furthermore, all the experiments are conducted under the Ubuntu system by using Pytorch deep learning framework, on a workstation with 4 GeForce GTX 1080Ti GPUs.

\subsection{Performance Analysis}
In this section, we present the performance analysis to evaluate the effectiveness of the proposed POA-OSL in four aspects, including the benchmark evaluations, the ablation study of different proposed components, the performance analysis of specific classes and the impacts of different parameter settings. For this purpose, the experiments are conducted on the widely used and well-known NTU RGB+D 60, NTU RGB+D 120 and PKU-MMD benchmark datasets.

\subsubsection{Benchmark Evaluations}

To verify the efficacy of the proposed POA-OSL, we perform the benchmark evaluations with two different dataset partitioning, which are medical action partitioning and general dataset partitioning. Quantitative experimental results compared with the recent state-of-the-art approaches are shown in Table \uppercase\expandafter{\romannumeral1} and Table \uppercase\expandafter{\romannumeral2}.  

\begin{table}[htbp!]
  \centering
  \caption{The 5-way-1-shot Top-1 accuracy (\%) comparisons with SOTA methods on NTU RGB+D 60, NTU 120 RGB+D and PKU-MMD datasets for medical action recognition.}
  \label{tab:pkummd_metrics}
  \begin{tabular}{|c|c|c|c|}
    \hline
    \multirow{2}{*}{\centering \textbf{Approaches}} & \multirow{2}{*}{\textbf{NTU-60}} & \multirow{2}{*}{\textbf{NTU-120}} & \multirow{2}{*}{\textbf{PKU-MMD}} \\
    & & & \\ \hline
    Skeleton-DML \cite{memmesheimer2022skeleton} & 45.1 & 39.3  & 34.8\\
    SL-DML \cite{memmesheimer2020signal} & 33.6 & 41.8 & 36.0 \\
    PAMMAR \cite{xie2023one} & 56.5 & 67.9 & 57.3 \\ \hline
    Position & 56.5 & 67.9 & 57.3 \\
    Orientation & 56.1 & 64.3 & 68.4 \\
    POA-OSL (MF) & \textbf{57.3} & \textbf{68.6} & \textbf{72.0}  \\
    POA-OSL (MS) & \textbf{59.2} & \textbf{68.6} & \textbf{68.6}  \\ \hline
  \end{tabular}
\end{table}

\begin{table}[htbp!]
\centering
\caption{The 5-way-1-shot human action recognition Top-1 accuracy (\%) comparisons with SOTA methods on NTU RGB+D 60, NTU RGB+D 120 and PKU-MMD with general dataset partitioning. $\dagger$ indicates the SIG method is applied.}
\vspace*{3mm}
\begin{tabular}{|c|c|c|c|} \hline
\textbf{Approaches} & \textbf{NTU-60} &\textbf{NTU-120} &\textbf{PKU-MMD}\\ \hline\hline
Attention Network \cite{liu2017global}& - & 41.0 & -\\
Fully Connected \cite{liu2017global}  & 60.9 & 42.1 & 56.4\\
Average Pooling \cite{liu2017skeleton} & 59.8 &  42.9 & 58.1\\
APSR \cite{liu2019ntu} & - & 45.3 & - \\
TCN \cite{Alb2021one} & 64.8 & 46.5 & 56.1\\
CTR-GCN-KP \cite{wang2023neural} & - & 68.1 & -\\
$\dagger$ SL-DML \cite{memmesheimer2020signal}& 71.4 & 50.9 & 67.0\\
ALCA-GCN \cite{zhu2023adaptive} & - & 57.6 & -\\
PartProtoNet\ \cite{chen2023part} & - & 65.6 & -\\
$\dagger$ Skeleton-DML \cite{memmesheimer2022skeleton} & 71.8 & 54.2 & 68.6\\ 
SMAM-Net \cite{li2022smam} & 73.6 & 56.4 & 70.4\\
$\dagger$ PAMMAR \cite{xie2023one} & 69.9 & 58.3 & 78.5\\
MMTS \cite{lee2023mmts} & - & 69.8 & -\\\hline\hline
$\dagger$ \textit{POA-OSL} & \textbf{76.3} & \textbf{76.0} & \textbf{82.6}\\ \hline
\end{tabular}
\end{table}


Table \uppercase\expandafter{\romannumeral1} reports the comparisons between the proposed POA-OSL and the other state-of-the-art one-shot learning methods on medical action recognition. Despite the single orientation feature does not achieve the best performance, our proposed POA-OSL achieves the best Top-1 accuracy regardless of different benchmark datasets compared with the other state-of-the-art methods, verifying that the orientation features can assist the position features in further improving the recognition performance by enhancing the discriminating abilities.

As could be observed from Table \uppercase\expandafter{\romannumeral2}, the proposed POA-OSL outperforms the other recent state-of-the-art one-shot learning approaches for human action recognition with general dataset partitioning, which indicates our proposed method has the capabilities to provide more informative and discriminate features for model training in all the three widely-used and well-known benchmark datasets. It is worth noting that both the proposed CsA and DTW modules are applied in the proposed POA-OSL framework for comparing with the state-of-the-art methods in this part of the experiments by the transformed signal-level images with 192$\times$192 resolutions.

\subsubsection{Ablation Study}

Table \uppercase\expandafter{\romannumeral3} reports the detailed ablation study for the proposed POA-OSL on the benchmark datasets, where w/o (without) CsA indicates only the DTW module is applied and w/ (with) CsA indicates both the CsA and DTW modules are employed. The performance of POA-OSL (MF) is improved from 55.0\% to 57.3\% on NTU-60 and from 66.6\% to 72.0\% on PKU-MMD by applying the proposed CsA module, which verifies the proposed CsA module could effectively reducing the misclassification issue between the similar actions by correctly guiding the model to focus on the important parts. 

Compared to our previous work PAMMAR \cite{xie2023one}, the proposed method here effectively reduces incorrect recognition and achieves improved performance. To be precise, compared with the single position-level feature, the proposed method improves the performance from 56.5\% to 59.2\% on NTU-60. Improved performance is also achieved on the NTU-120 and PKU-MMD datasets, which achieve 68.6\% and 72.0\% Top-1 accuracy, respectively.

\begin{table}[htbp!]
  \centering
  \caption{The ablation study of the proposed POA-OSL for medical action recognition on different benchmark datasets. The 5-way-1-shot Top-1 accuracy (\%) results are provided.}
  \label{tab:ntu120_metrics}
  \scalebox{0.81}
  {
  \begin{tabular}{|c|cc|cc|cc|}
    \hline
    \multirow{2}{*}{\centering \textbf{Features}} & \multicolumn{2}{c|}{\textbf{NTU-60}} & \multicolumn{2}{c|}{\textbf{NTU-120}} & \multicolumn{2}{c|}{\textbf{PKU-MMD}} \\
    \cline{2-7}
    & \textbf{w/o CsA} & \textbf{w/ CsA} & \textbf{w/o CsA} & \textbf{w/ CsA} & \textbf{w/o CsA} & \textbf{w/ CsA} \\
    \hline
    Position \cite{xie2023one} & 52.9 & 56.5 & 64.6 & 67.9 & 56.4 & 57.3 \\
    Orientation & 52.1 & 56.1 & 63.0 & 64.3 & 65.2 & 68.4 \\
    POA-OSL (MF) & \textbf{55.0} & \textbf{57.3} & \textbf{67.8} & \textbf{68.6} & \textbf{66.6} & \textbf{72.0}\\ 
    POA-OSL (MS)  & \textbf{58.2} & \textbf{59.2} & \textbf{67.1} & \textbf{68.6} & \textbf{65.4}& \textbf{68.6}\\ \hline
  \end{tabular}
}
\end{table}

The visualization heatmap comparisons between w/o CsA and w/ CsA for staggering and headache are illustrated in Fig. 7. For the A042 staggering, both the arms and legs of the subjects keep significant movements in the sequences. It could be observed that the model correctly focuses on the important body parts (leg and foot) after the CsA module with the position features, which are labelled as 15, 19, 14 and 18. For the A044 headache, the subjects keep holding their heads by using their arms. The heatmaps for A044 demonstrate higher weights after getting through the CsA module for the arms and hands in the position features, which are labelled as 6, 10, 11, 7, 21, 22, 23 and 24. Moreover, the weights of the head are slightly improved in the orientation features after the CsA module which corresponds to the minor movements of the head, which are labelled as 2 and 3.  Detailed illustrations of the landmark labels are shown in Fig. 4. Comparing the w/o CsA heatmap results between the position and orientation for A042, it could be observed that landmarks 6, 7, 10 and 11 have more important weights in orientation features than position features. This situation also happens in A044, in which the color of these parts in the orientation features is brighter than in the position features. The reason is that the orientation features contain more valuable information to interpret the actions. The aforementioned visualization results verify the complementary relation between the position and orientation-level features. Moreover, they demonstrate the CsA module could effectively guide the network to focus on the important information for each specific medical action, which is beneficial for distinguishing similar actions and further improving the model performance.

Moreover, the UMAP \cite{mcinnes2018umap} visualization results of medical actions from the NTU RGB+D 120 dataset are shown in Fig. 8. The uniform manifold approximation and projection (UMAP) demonstrate the clustering relationship by dimension reduction operation of the features. The larger distance from the other clusters demonstrates higher performance. For example, compared with the single position or orientation features, the falling down is more centralized with the proposed method, which is illustrated on the left of Fig. 8.

\subsubsection{Analysis for Specific Classes}

\begin{figure*}[htbp!]
\centering
\includegraphics[width=16.1cm, height=7.6cm]{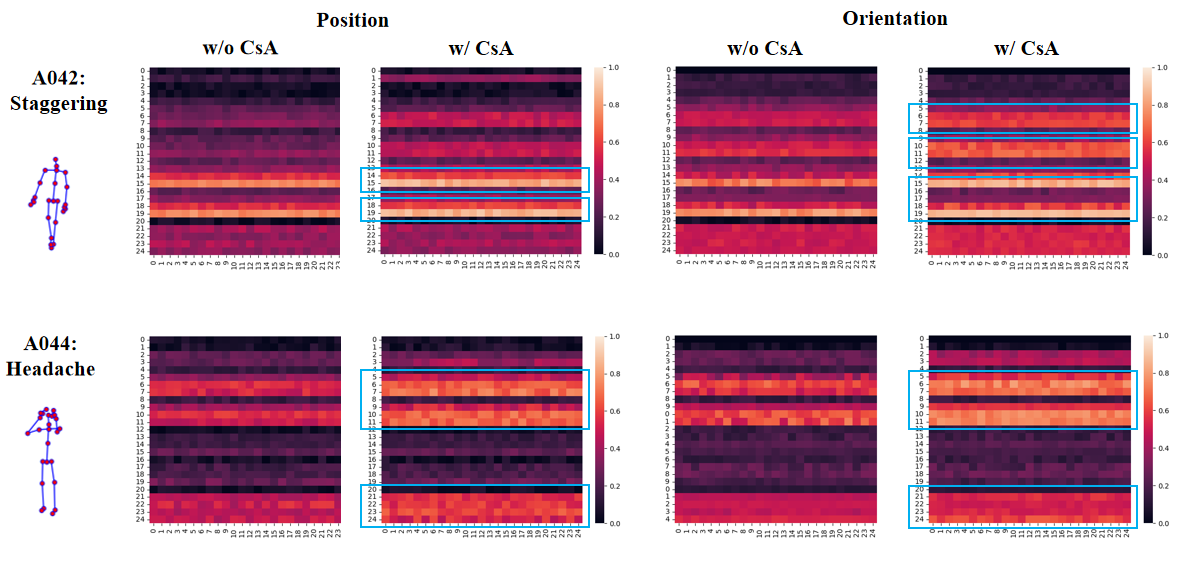}
\caption{The heatmap visualization for 5-way-1-shot medical action recognition implemented by our proposed POA-OSL on A042 (staggering) and A044 (headache) actions from NTU RGB+D 120 dataset after processing the DTW and CsA modules. We highlight the predicted important body parts in the blue boxes (best viewed in the color version).}
\centering
\end{figure*}

\begin{figure*}[htbp!]
\centering
\includegraphics[width=17.1cm, height=5.4cm]{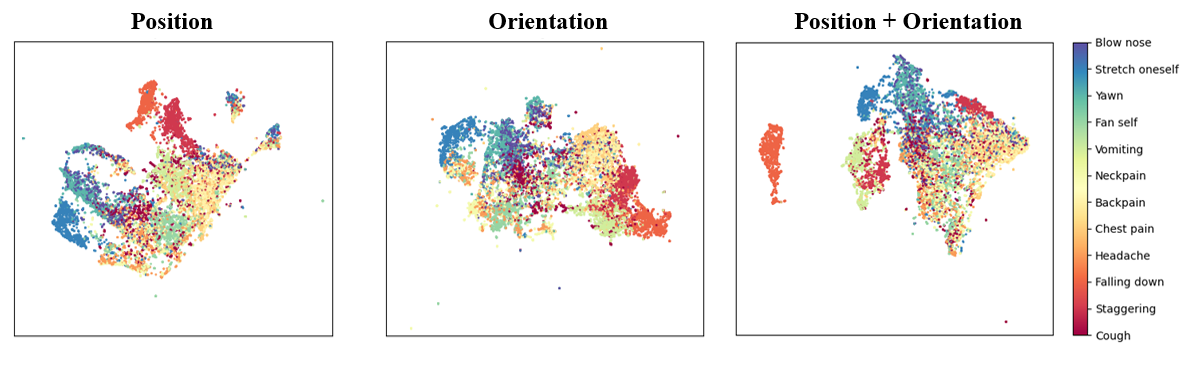}
\caption{The UMAP \cite{mcinnes2018umap} visualization for 5-way-1-shot performance implemented by the proposed POA-OSL on NTU-RGB+D 120 dataset with different features, which are position features (right), orientation features (middle) and position+orientation features (left). A more concentrated cluster indicates the action with higher performance. (best viewed in the color version).}
\centering
\end{figure*}

To analyze the performance of each specific action, we conduct quantitative ablation study experiments with 192$\times$192 resolutions for each specific medical action on the benchmark datasets by applying the proposed POA-OSL in this part. 

Table \uppercase\expandafter{\romannumeral4} reports the Top 1 accuracy (\%) performance on the medical action for the three datasets. To compare the performance of orientation with position features in Table \uppercase\expandafter{\romannumeral4}, for some specific medical actions, such as vomiting and headache in NTU RGB+D 120, backache and heart pain from PKU-MMD dataset, the performance of single orientation feature relatively outperforms single position feature performance. This is due to the fact that the variation in the orientations of the human physical landmarks is more significant in these medical action sequences, which contributes informative knowledge for model training.

\begin{table*}[htbp!]
  \centering
  \caption{Ablation study on the specific classes on NTU RGB+D 60 (NTU-60), NTU RGB+D 120 (NTU-120) and PKU-MMD datasets for 5-way-1-shot medical action recognition with Top 1 Accuracy (\%). The w/o CsA indicates the model only contains the DTW module and w/ CsA indicates the model contains both DTW and CsA modules.}
  \label{tab:specific class acc}
  \begin{tabular}{|c|c|c|c|c|c|c|c|c|c|c|}
    \hline
    \multirow{2}{*}{\centering \textbf{\#}} & \multirow{2}{*}{\centering \textbf{Medical Action}} & \multirow{2}{*}{\centering \textbf{Datasets}} &\multicolumn{2}{c|}{\textbf{Position}} & \multicolumn{2}{c|}{\textbf{Orientation}} & \multicolumn{2}{c|}{\textbf{POA-OSL (MF)}} & \multicolumn{2}{c|}{\textbf{POA-OSL (MS)}} \\
    \cline{4-11}
    & & & \textbf{w/o CsA} & \textbf{w/ CsA} & \textbf{w/o CsA} & \textbf{w/ CsA } & \textbf{w/o CsA} & \textbf{w/ CsA} & \textbf{w/o CsA} & \textbf{w/ CsA} \\ \hline\hline
    A41 & Cough & NTU-60 & 33.2 & 42.5 & 31.4 & 43.1 & 34.2 & 40.2 & 43.1 & \textbf{45.8}\\
    A43 & Falling & NTU-60 & 97.7 & 98.2 & 95.0 & 95.8 & \textbf{99.4} & 97.8 & 98.6 & 99.1\\
    A44 & Headache & NTU-60 & 62.5 & 57.4 & 55.3 & 57.1 & 57.7 & 57.7 &  63.4 & \textbf{63.5}\\ 
    A45 & Chest pain & NTU-60 & 32.9 & 45.9 & 32.4 & 44.9 & 37.3 & 47.5 & 46.0 & \textbf{48.1} \\
    A46 & Back pain & NTU-60 & 49.8 & 53.6 & 48.0 & 54.7 & 59.1 & \textbf{60.2} &  53.1 & 56.9\\ 
    A47 & Neck pain & NTU-60 & 42.9 & 42.8 & 41.3 & 41.6 & 43.1 & 43.0 &  42.1 & \textbf{45.2}\\ \hline \hline
    A41 & Cough & NTU-120 & 36.5 & 37.5 & 42.9 & 43.0 & 43.0 & 43.5 & 45.0 & \textbf{45.6}\\
    A42 & Staggering & NTU-120 & 88.4 & 92.2 & 91.0 & 93.1 & 90.3 & 93.0 & 93.0 & \textbf{95.7}\\
    A43 & Falling & NTU-120 & 89.2 & 95.7 & 89.0 & 94.8 & 98.6 & 98.9 & 95.0 & \textbf{99.5}\\
    A44 & Headache & NTU-120 & 61.8 & 63.1 & 64.1 & 66.5 & \textbf{70.2} & 66.1 & 66.8 & 65.7\\ 
    A45 & Chest pain & NTU-120 & 48.9 & 50.4 & 49.4 & 50.4 & 51.9 & \textbf{54.8} & 53.9 & 54.1\\ 
    A46 & Back pain & NTU-120 & 62.6 & 65.1 & 61.2 & 63.4 & 64.7 & \textbf{68.8} & 65.7 & 67.8\\ 
    A47 & Neck pain & NTU-120 & 45.3 & 54.9 & 46.3 & 49.8 & 53.7 & 53.7 & 48.0 & \textbf{55.9}\\
    A48 & Vomiting & NTU-120 & 65.3 & 63.6 & 70.9 & 72.3 & 60.8 & \textbf{73.2} & 72.5 & 73.0\\
    A49 & Fan self & NTU-120 & 62.4 & 73.2 & 53.9 & 63.9 & 68.4 & \textbf{75.2} & 62.2 & 72.6\\
    A103 & Yawn & NTU-120 & 60.5 & 68.1 & 67.1 & 67.4 & \textbf{75.4} & 69.3 & 69.8 & 74.4\\
    A104 & Stretch oneself & NTU-120 & 72.5 & 78.9 & 66.6 & 69.3 & \textbf{82.3} &  74.2 & 72.4 & 79.0\\
    A105 & Blow nose & NTU-120 & 58.2 & 59.8 & 59.5 & 56.0 & \textbf{64.5} & 61.7 & 63.6 & 60.4\\ \hline \hline
    A11 & Falling & PKU-MMD & 98.5 & 99.6 & 96.6 & 97.5 & 93.8 & 97.7 & 99.4& \textbf{99.8}\\
    A42 & Backache & PKU-MMD & 49.6 & 49.6 & 60.0 & \textbf{75.3} & 72.1 & 72.2 & 63.8 & 74.7\\
    A43 & Heart pain & PKU-MMD & 42.8 & 43.4 & 65.7 & 65.9 & 56.0 & \textbf{67.7} & 60.1 & 61.9\\
    A44 & Headache & PKU-MMD & 47.0 & 50.1 & 59.1 & 60.8 & 60.9 & \textbf{69.7} & 56.1 & 58.1\\
    A45 & Neck pain & PKU-MMD & 44.5 & 46.6 & 46.3 & 45.6 & 45.8 & \textbf{51.0} & 47.8 & 48.4\\ \hline 
  \end{tabular}
\end{table*}

\begin{figure*}[htbp!]
  \centering
  \resizebox{0.9\textwidth}{!}{
  \begin{minipage}[b]{0.36\textwidth}
    \centering
    \begin{subfigure}{\textwidth}
      \centering
      \begin{tikzpicture}
          \draw[dashed] (0,0) grid (5,4); 
          
          \foreach \y/\ylabel in {0/56, 1/57, 2/58, 3/59, 4/60} {
            \node[left] at (0,\y) {\ylabel};
          }
          \node[left, rotate=90] at (-0.8,3.5) {Top 1 accuracy (\%)};
          
          \foreach \x/\xlabel in {0/0, 1/0.4, 2/0.8, 3/1.2, 4/1.6, 5/2.0} {
            \node[below] at (\x,0) {\xlabel};
          }
          \node[below] at (2.5,-0.4) {hyperparameter $\alpha$ value};
          
          \coordinate (lastpoint) at (0,0.50);
          \foreach \x/\y in {0/0.50, 1/2.64, 2/3.14, 3/3.20, 4/3.07, 5/2.91} {
            \draw[fill=red, circle, thick] (\x,\y) circle (3pt);
            \draw[fill=red, thick] (\x,\y) -- (lastpoint);
            \coordinate (lastpoint) at (\x,\y);
          }
          \node[above] at (2.5,4.1) {NTU RGB+D 60};
    \end{tikzpicture}
      \label{fig:sub1}
    \end{subfigure}
  \end{minipage}
  \hfill
  \begin{minipage}[b]{0.4\textwidth}
    \centering
    \begin{subfigure}{\textwidth}
      \centering
          \begin{tikzpicture}
          \draw[dashed] (0,0) grid (5,4); 
          
          \foreach \y/\ylabel in {0/65, 1/66, 2/67, 3/68, 4/69} {
            \node[left] at (0,\y) {\ylabel};
          }
          
          \foreach \x/\xlabel in {0/0, 1/0.4, 2/0.8, 3/1.2, 4/1.6, 5/2.0} {
            \node[below] at (\x,0) {\xlabel};
          }
          \node[below] at (2.5,-0.4) {hyperparameter $\alpha$ value};
          
          \coordinate (lastpoint) at (0,2.86);
          \foreach \x/\y in {0/2.86, 1/3.42, 2/3.61, 3/3.50, 4/3.21, 5/2.82} {
            \draw[fill=blue, circle, thick] (\x,\y) circle (3pt);
            \draw[thick] (\x,\y) -- (lastpoint);
            \coordinate (lastpoint) at (\x,\y);
          }
          \node[above] at (2.5,4.1) {NTU RGB+D 120};
        \end{tikzpicture}
      \label{fig:sub2}
    \end{subfigure}
  \end{minipage}
  \hfill
  \begin{minipage}[b]{0.36\textwidth}
    \centering
    \begin{subfigure}{\textwidth}
      \centering
      \begin{tikzpicture}
          \draw[dashed] (0,0) grid (5,4); 
          
          \foreach \y/\ylabel in {0/57, 1/61, 2/65, 3/69, 4/73} {
            \node[left] at (0,\y) {\ylabel};
          }
          
          \foreach \x/\xlabel in {0/0, 1/0.4, 2/0.8, 3/1.2, 4/1.6, 5/2.0} {
            \node[below] at (\x,0) {\xlabel};
          }
          \node[below] at (2.5,-0.4) {hyperparameter $\alpha$ value};
          
          \coordinate (lastpoint) at (0,0.30);
          \foreach \x/\y in {0/0.30, 1/2.27, 2/2.8, 3/3, 4/3.34, 5/3.48} {
            \draw[fill=green, circle, thick] (\x,\y) circle (3pt);
            \draw[thick] (\x,\y) -- (lastpoint);
            \coordinate (lastpoint) at (\x,\y);
          }
          \node[above] at (2.5,4.1) {PKU-MMD};
        \end{tikzpicture}
      \label{fig:sub3}
    \end{subfigure}
  \end{minipage}
  }
  \caption{Comparisons of Top 1 accuracy (\%) with different $\alpha$ hyperparameter settings on NTU RGB+D 60, NTU RGB+D 120 and PKU-MMD datasets. The ablation study is computed under the full model, which contains both the CsA and DTW modules. The resolution sizes are set as 192$\times$192.}
  \label{fig:combined}
\end{figure*}
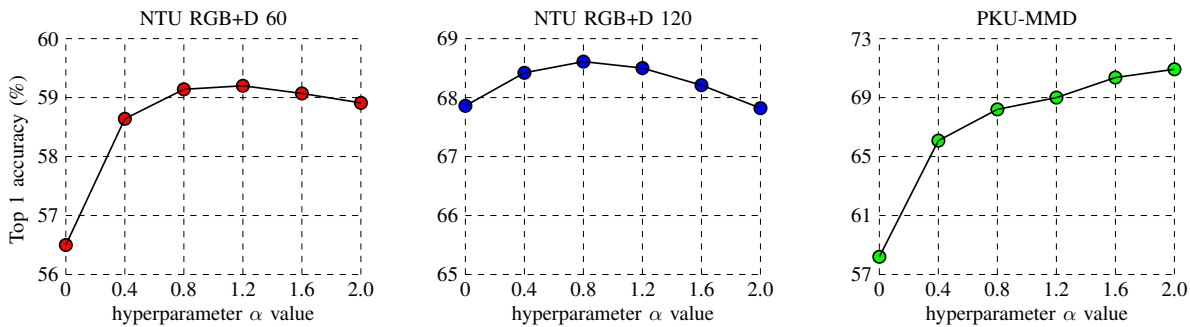

Overall, our proposed POA-OSL achieve the best performance among almost all the medical actions compared to the single feature performance in all the datasets. This verifies that the position and orientation features are complementary in the spatial and temporal dimensions. Position features provide the localization of the landmarks and orientation features provide the bending magnitude of the subject bodies. Both the changing rates of these two features are also supplied in the temporal dimension. Compared to \cite{xie2023one}, the proposed assisted training method achieves 7.1\% and 3.7\% improved performance for headache and back pain on NTU RGB+D 120, respectively. Similarly, the proposed method outperforms the position features on vomiting by 9.6\%. Moreover, by applying the proposed CsA mechanism and orientation assisted training method, the accuracy of neck pain from NTU-120 is significantly improved from 45.3\% to 55.9\%, which validates the hypothesis previously illustrated in Fig. 2. Furthermore, according to the results from the three datasets, particularly in PKU-MMD, falling achieved over 99\% accuracy. This is because both the position and orientation of falling have noticeable movements based on the landmark localization and body bending angles. There exist significant differences between the first frame and the last frame for falling. The above experimental evaluations imply that our proposed method could considerably strengthen the performance of medical action recognition. 

As we can see from Table \uppercase\expandafter{\romannumeral4}, both cough and chest pain are the most difficult medical actions to recognize. The reason for this is that both the movements of these two medical actions are slight in both spatial and temporal dimensions. In contrast, the staggering and falling achieve the most and the second-most promising accuracy on the NTU-120 dataset, which are 95.7\% and 99.5\%, respectively. It could be observed that the performance of headache from PKU-MMD is remarkably improved from 50.1\% to 69.7\% by applying the proposed POA-OSL (MF), which further verifies that POA-OSL could enhance the discriminating ability on different datasets.

The suggestions for future work can be attributed to two aspects, one is to exploit the feature relation between position and orientation, which is useful for capturing the long-term latency of the trajectories of important body parts to further improve the performance of the proposed method. The other one is that the medical actions with slight variations in motion should be given more attention since recognizing these actions is relatively challenging, and it implies that improving the accuracy of recognizing these medical actions can enhance the performance of the overall framework.

\subsubsection{Parameters Analysis}

We also conduct experiments on the datasets to analyze the impacts of different critical parameter settings. The proposed approach is evaluated on different transformed image resolutions, as illustrated in Table \uppercase\expandafter{\romannumeral5}. Top 1 accuracy (\%) is employed to investigate the relative influences on the performance. With the increase of resolutions from 32 to 192, the performance of the proposed approach is significantly improved from 58.9\% to 67.9\%. This is because there exists a positive correlation between the image resolution and the performance. Since the images contain more information with larger image resolutions.

\begin{table}[htbp!]
\centering
\caption{The 5-way-1-shot Top-1 accuracy (\%) of the proposed method using on different signal image resolutions for medical actions on NTU RGB+D 120 dataset.}
\begin{tabular}{|c|c|cc|} \hline
\multirow{2}{*}{\textbf{Resolutions}} & \multirow{2}{*}{\textbf{Baseline} \cite{memmesheimer2022skeleton}}  & \multirow{2}{*}{\textbf{DTW}} & \multirow{2}{*}{\textbf{CsA+DTW}}\\
& & & \\\hline 
32$\times$32 & 33.4 & 56.0 &\textbf{58.9} \\
64$\times$64 & 33.7 & 60.7 &\textbf{62.0} \\
96$\times$96 & 37.6 & 59.5 &\textbf{64.7} \\
144$\times$144 & 40.7 & 63.5&\textbf{65.4} \\
160$\times$160 & 42.0 & 64.2 &\textbf{66.0} \\
192$\times$192 & 41.8 & 64.6 &\textbf{67.9} \\ \hline
\end{tabular}
\end{table}

Moreover, we analyze the influence of different $\alpha$ value settings on the medical action recognition performance, which control the importance of orientation features for assisting in improving the classification performance. The experimental results of different importance are shown in Fig. 9. Since both the position and orientation images contain different action features in different dimensions. In order to determine the best weight parameter settings for different datasets, we experimentally determine an appropriate value for $\alpha$ from (8) to manage the importance of the weight between the position features and orientation features. To this end, we conducted a set of pilot tests $\alpha = \{0, 0.4, 0.8, 1.2, 1.6, 2.0\}$ on NTU RGB+D 60, NTU RGB+D 120 and PKU-MMD datasets. All the performance of the three datasets increases when the value of $\alpha$ changes from 0 to 0.4. The best important weight selections for NTU RGB+D 60, NTU RGB+D 120 and PKU-MMD are 1.2, 0.8 and 2.0, respectively. The above experimental results demonstrate that the position and the orientation features are complementary for enhancing the medical action recognition performance. One thing that needs to be noted is that we use the largest image resolutions with the full proposed model and keep them fixed during this ablation study experiment.

\subsection{Failure Case}
According to Table \uppercase\expandafter{\romannumeral4}, our POA-OSL performs better in recognizing some of the medical actions rather than the baseline approach \cite{xie2023one}, such as "Falling down" and "Staggering", demonstrating that our method of enhancing the complementary relation features in distinguishing different medical actions. However, the specific experimental results of the action "Cough" are relatively not promising, which are around 45\%. This demonstrates our method is insensitive in capturing the medical actions with the tiny range of movements. Presumably, this is because the pixel values of the orientation features for these medical actions are relatively limited. Moreover, since the backbone is chosen to use ResNet18, the kernel size of it is fixed when taking the instances, the feature values may extend to disappear after several layers compared to the other actions. Therefore, the discriminative capability of POA-OSL for some challenging medical actions with slight movements is insufficient.


\section{Conclusion}
In this paper, we proposed a position and orientation-aware one-shot learning framework which contains dynamic temporal warping and cross-attention modules to improve the medical action recognition performance, which is named POA-OSL. The proposed method aims to address similar action, temporal mismatching, privacy protection and medical data limitation issues. The effectiveness of each module from the proposed framework is analyzed in detail. Extensive experimental results demonstrate that our proposed one-shot learning approach achieves the best accuracy performance compared with the baseline approaches on NTU RGB+D 120, NTU RGB+D 60 and PKU-MMD datasets for medical action recognition. General partitioning on the most challenging NTU RGB+D 120 achieved 76.0\% Top-1 accuracy for 5-way-1-shot learning, which outperformed the other state-of-the-art methods by 6.2\%. The proposed method significantly improves performance and is suitable for human abnormal activity detection in healthcare and medical-related applications.


\bibliographystyle{IEEEtran}
\bibliography{LeiyuXie}
\end{document}